\documentclass[runningheads]{llncs}

\usepackage{graphicx}
\usepackage{amsmath}
\usepackage{listings}
\usepackage{amsmath}
\usepackage{algorithm}
\usepackage[noend]{algpseudocode}
\usepackage{csquotes}
\usepackage{fancyhdr}

\begin{document}
\title{``Why did you do that?''}
\subtitle{Explaining black box models with Inductive Synthesis}
\author{G{\"o}rkem Pa{\c{c}}ac{\i}
\and David Johnson
\and Steve McKeever
\and Andreas Hamfelt
}
\authorrunning{Pa{\c{c}}ac{\i} et al.}
\institute{Department of Informatics and Media, Uppsala University,\\
Box 513, 75120 Uppsala, Sweden\\}

\maketitle  

\begin{abstract}
By their nature, the composition of black box models is opaque. This makes the ability to generate explanations for the response to stimuli challenging. The importance of explaining black box models has become increasingly important given the prevalence of AI and ML systems and the need to build legal and regulatory frameworks around them. Such explanations can also increase trust in these uncertain systems. In our paper we present RICE, a method for generating explanations of the behaviour of black box models by (1) probing a model to extract model output examples using sensitivity analysis; (2) applying \textit{CNPInduce}, a method for inductive logic program synthesis, to generate logic programs based on critical input-output pairs; and (3) interpreting the target program as a human-readable explanation. We demonstrate the application of our method by generating explanations of an artificial neural network trained to follow simple traffic rules in a hypothetical self-driving car simulation. We conclude with a discussion on the scalability and usability of our approach and its potential applications to explanation-critical scenarios.

\keywords{artificial intelligence, machine learning, black box models, explanation, inductive logic, program synthesis}
\end{abstract}

\section{Introduction}
\label{intro}
Much of the software industry is characterised by a clear separation of concerns, encouraging users not to be attentive to  how a system functions. However with safety critical applications or those that must adhere to tight legal requirements, further assurances are needed. Rigorous software engineering, through formal development or comprehensive testing, has been
very successful at supporting complex system development in many exacting fields. However we are now moving into an era where software is generated from examples.

The concept of the \textit{black box} describes a system that can only be viewed with reference to its observed inputs and resulting outputs. Black box models can be validated using well-established software testing methods based on such observations \cite{Beizer:1995:BTT:202699}. This validates the intended behaviour of a black box model by providing us an \textit{interpretation} of the mechanics of the model. Why the observed behaviour occurs is opaque and any \textit{explanation} for its behaviour remains uncertain.

AI and Machine Learning (ML) models are often black boxes. While there are some ML models that exhibit simple relationships that can be straightforwardly explained (e.g. logistic regression) or that inherently display logical structure (e.g. decision trees), complex models that consist of artificial neural networks (ANNs), deep neural networks (DNNs) and random forests are non-intuitive and are not structured with human-readable logic. This makes us unable to posit the question to a model, \textit{``Why did you do that?''}, to seek explanations of the decisions that ANN-based systems make \cite{Gunning2018}.

To provide explanations of black box models is an intersection of technical and human challenges:
\begin{itemize}
    \item How can we generate an explainable model of a black box? 
    \item How can we ensure these explainable models are human-readable?
    \item How do we use these explanations of black boxes?
\end{itemize}

\textbf{Contributions:} In this paper we present a novel method, Rule Induction of CNP Explanations (RICE), that combines sensitivity analysis and logic program synthesis for generating explanation models of black boxes. The explanation models are not intended to be executable, but rather to be human-readable in their explanation of the executable model. Previous work on explaining predictions of ML classifiers, such as the LIME algorithm \cite{Ribeiro:2016:WIT:2939672.2939778}, which uses a local linear approximation of a model's behaviour to highlight what a model has picked out in the input data to make its classification. Such work focuses on explaining classifiers, while our method focuses on explaining the rules and logic learned in a black box. 

The paper is structured as follows. In Section~\ref{related} we position our work within the field of black box interpretability. In Section~\ref{explanation} we provide an overview of the RICE methodology. In Section~\ref{demonstration} we go through a simple traffic light use case, showing how the explainable model is produced using RICE. In Section~\ref{discussion} we reflect on our methodology and in Section~\ref{conc} we summarise and list some areas of future work.




\section{Related work}
\label{related}
Black box models, and in particular ML models, now exist in many critical domains such as medicine, finance, and even in military applications. There is a clear problem faced by the industrialization of using such models, where the human understanding of how they behave and work is difficult. Model \textit{interpretability} sets out to produce metrics about models, for example, scoring the performance of predictions against ground truth, and to build trust in models by providing examples of when they perform correctly or otherwise \cite{DBLP:journals/cacm/Lipton18}. When applied to black box models, for example in supervised learning models, such measures are focused on validating the learning of associations between factors. This of course does not guarantee that the model has learned causal relationships. Associations and causality describe the behaviour of models, however, they do not reveal an \textit{explanation} for why an output was generated nor an explanation for how the black box works.

Interpretable models already exist. Of note, there are a number of classification models that are readily interpretable, for example decision trees, rule-based systems, and nearest-neighbour methods, each with varying levels of usability \cite{kar38534}. Decision trees generate a tree-like structure representing a series of tests on different features in a training dataset where leaf nodes represent various labeled classifications. Rules-based systems explicitly map an input to an action through some explicitly defined series of logical assertions (\textit{if-then} rules). Nearest-neighbour algorithms qualify a classification based on the values of attributes in the immediate neighbourhood of some input. 

In \cite{Guidotti:2018:SME:3271482.3236009} the authors produced an ``Open the black box taxonomy'' to better understand different approaches to explaining black box models. At the top level there are two options: To formulate a black box explanation, and design a transparent box. The latter has been discussed in some depth in \cite{DBLP:journals/corr/abs-1811-10154} where its author argues that explanations are often unreliable and misleading, and therefore black box models should not be used for critical systems. She cites several examples where lack of transparency and accountability of predictive models have had severe consequences. Counter-arguments to interpretable ML models include black boxes holding significant value and intellectual property for the model authors, and that interpretable models can be more costly to construct. 
They further define the \textit{black box explanation problem} that requires an explanation model that is ``globally interpretable''; that is to be able to mimic the behaviour of the black box \textit{and} should be understandable by a human.

Approaches to generating human-readable explanation models include the Automatic Statistician, a system that discovers plausible models from data and automatically presents its findings as figures and as natural language \cite{Ghahramani:2015:Nature:26017444,Lloyd:2014:ACN:2893873.2894066},  LIME (Local Interpretable Model-agnostic Explanations) \cite{Ribeiro:2016:WIT:2939672.2939778}, a method for isolating parts of a given input that contribute most to a classification, and QII (Quantitative Input Influence) \cite{7546525}, a technique for calculating influences of individual inputs or groups of inputs, each provides explanations as summaries of local causal phenomena. A recent survey lists various methods in explainable AI, and notes that due to rise in autonomous systems and complex models, there is even more need for interpretable models \cite{biran2017explanation}. 

\section{Explanations through Inductive Synthesis}
\label{explanation}
With RICE, we propose a method that combines sensitivity analysis and inductive logic programming. In contrast to LIME and QII, our approach seeks to generate globally interpretable model explanations for rule-based black box by synthesizing human-readable logic programs.

\begin{algorithm}
\caption{Rule Induced CNP Explanation (RICE)}\label{methodAlgorithm}
\begin{algorithmic}[1]
\Procedure{RICE}{training data}
\State $\textit{model} \gets \text{train}(\text{training data})$
\State $observables \gets $\textit{probe}(model)
\State $explanation \gets $\textit{CNPInduce}(observables)
\EndProcedure
\end{algorithmic}
\end{algorithm}

We start with a ML artefact after the learning step is accomplished. Sensitivity analysis consists of a series of methods developed to inspect the input/output relation of a function-like structure and is used in relation to ANNs to identify which input variables are relevant to the model, and which are not. This in turn is used to optimize the learning by reducing the unused dimensions in the training data. \cite{Zurada1994SensitivityAnalysis}. We use monothetic analysis, that is, systematically trying different values for one of the inputs while keeping the values for other inputs constant \cite{ten_broeke2016}. This is undertaken for each possible assignment of inputs to the model, effectively calculating every partial differential of the model's function. The data points where the partial differential $m$ is greater than an arbitrary $e$ are noted as \textit{critical examples}. Our method relies on the assumption that these extracted critical examples are a good estimate of the model's function. We call this stage of the method \textit{probing}.

For synthesizing logic programs we rely on earlier work on inductive program synthesis, CNPInduce (Induction of CNP) \cite{Pacaci2017CNP,Pacaci2017Representation}. The approach is a meta-interpretative form of inductive logic programming, where the CNPInduce synthesizer is written in Prolog. A meta-interpreter is a higher-level program that executes a program in a language and produces its input/output relation. In RICE, the meta-interpreter of a specially designed target language (CNP) is reversed, and this reversed meta-interpreter is executed with some known examples from its input/output relation thus producing all programs that would produce an input/output relation containing those examples \cite{HamfeltJFN94InductiveMetalogic,HamfeltJFN99InductiveSynthesis}. There are other rule-extraction methods that are applicable but they are mostly domain-specific, and require a user strategy to investigate the model. The strength of CNPInduce is its domain-agnostic technique and its ability to synthesize recursive programs \cite{Pacaci2017CNP,Pacaci2017Representation}. The ability to synthesize recursive programs makes it a good candidate to tackle models that deal with vector and matrix data such as audio and video signals. Even though inductive synthesis is a form of ML itself, it cannot perform well with noisy, high volume data; but it can produce a human-readable output. When coupled with a technique such as ANNs, which can deal with noisy input data, the combination gives a novel technique that has the efficiency of ANNs and the human-readable output of program synthesis. 

To combine the two techniques, we take the critical examples extracted from the black box model and export them in a format that can be input to CNPInduce, the synthesizer to generate CNP programs. By using critical examples the synthesizer generates a program in the CNP language which satisfies these observables, which constitutes our explanation model. We call this stage of the method simply \textit{synthesis}. Algorithm \ref{methodAlgorithm} illustrates this sequence of stages to generate explanations using RICE. Since the synthesized program and the ML artefact are semantically correlated through the critical examples, one can interpret the synthesized program instead of the ML artefact. This allows the inspection and validation of the ML model through the synthesized program. In RICE, this program is expressed in a language specifically developed for this purpose, called Combilog with Named Projection or CNP \cite{Pacaci2017Representation}. Its human-readability is improved compared to other forms of variable-free relational programming languages. Moreover, since it is a pure relational language the programs in this language can be automatically translated to more familiar languages, such as first-order logic, definite clauses \cite{Hamfelt1998combinatoryFormOfPure}, or even structured English \cite{Fuchs1996AttemptoControlledEnglish}. We call this stage of the method \textit{validation}. In the next section we demonstrate the RICE method through an example.

\section{A demonstration of model validation}
\label{demonstration}

To demonstrate our method we devised an experiment based on a simplification of a self-driving vehicle's decision system. We assume a case where through its sensors the vehicle's systems have identified the status of a traffic light (red, amber, green) and the distance to it. The vehicle needs to decide to accelerate or to brake continuously. Existing studies using ANNs on traffic lights focus on detection of the state of the traffic light under complex circumstances, and recent work demonstrate this is achievable in real-time \cite{BehrendtNovak2017Bosch}. An action stage which would naturally follow detection is usually left out of the model. For demonstration purposes of RICE we design a model that mimics the action stage. This allows us to focus on the core of the method instead of issues regarding optimization and scale. Let us assume this decision is left to an ANN, the model is trained with data from actual driving sessions, and it seems to function normally. In order to quantify the decision system's reliability, one may want to know precisely how it reacts to specific conditions. In the case of a trained model this is near-impossible since the model consists of binary data. Let us take it from here and show how our method can help us approach validation of this model. The implementation can be accessed in our GitHub repository\footnote{\url{https://github.com/UppsalaIM/rice/releases/tag/iccs19}}. We present this demonstration in multiple stages:

\begin{enumerate}
\item Training of the model
\item Extract examples from the model (probing)
\item Synthesize a human-readable program from examples (synthesis)
\item Inspect the synthesized program as an indirect representation of source artefact (validation).
\end{enumerate}

\subsection{Training of the model}
In our feature vector we encode each light as being on or off with 0 or 1, where position 0 corresponds to the red light, position 1 to the amber light and position 2 to the green light. Our final feature representing the distance of the car from the traffic lights is encoded as a floating point decimal between 0 and 1 representing the range of distances from 0m to 100m, and is placed at position 3 of the feature vector. Our sample states are therefore:
\begin{verbatim}
[1, 0, 0, 0.25]  # red and the car is 25m from the lights
[0, 1, 0, 1.0]   # amber and the car is 100m from the lights
[0, 0, 1, 0.0]   # green and the car is 0m from the lights
\end{verbatim}
A classification model is implemented and trained using TensorFlow\footnote{\url{https://tensorflow.org}}. The network has four layers, the input layer with 4 nodes, two hidden layers with 11 nodes each, and an output layer with 1 node. In the output layer, the value of the single output node represents the action; 1 for \textit{go}, and 0 for \textit{stop}. 

Training data was procedurally generated in order to be able to embed algorithmic rules in the training data. While calculating the training labels a 2\% noise was introduced. This is important to show that the method works with noisy data, as inductive synthesis methods by themselves would not be able to work with noisy examples.
\label{subs_generate_and_train}
Training data is generated with random values for the state of the traffic light and the distance, within the valid space of input vectors. The labels are calculated according to the following scheme:

\begin{itemize}
\item If the light is \textbf{red}, and the distance to it is less than 60m, then stop; otherwise go. For example, if input is $[1,0,0,0.5]$, output is $0$; and if the input is $[1,0,0,0.9]$, output is $1$.
\item If the light is \textbf{amber}, and the distance to it is between 10m and 80m, then stop; otherwise go. The minimum distance of 10m is for avoiding stopping when it is too close. For example, if input is $[0,1,0,0.2]$, output is $0$.
\item If the light is \textbf{green}, go. For example, if input is $[0,0,1,0.1]$, output is $1$.
\end{itemize}

In the training set there were 45,000 samples, and in the testing set there were 5,000. The data was generated as 50,000 samples and then split into training versus test data. During 10 training runs with 10 epochs each, accuracy up to 99.7\% was measured, and it was consistently above 96\%. 

A model with 99\% accuracy was saved to a HDF5 file which contains the model structure and the weights. In the following stage \textit{probing}, we load this model from a file in order to show that the training data is separate from the next stage.

\subsection{Probing: Extracting examples from a trained model}
The trained model from the previous stage is loaded from a file using TensorFlow with the probing stage implemented in Python. It is evaluated for each combination of the light state, for each possible distance (with a 1/100 granularity). The \verb|distance_sweep| function below displays how the input \textit{sweep} is generated, and the following two lines display how the model is evaluated for the given inputs. Here only the sweep for the red light state is displayed, where the outputs from the model are stored in the variable \verb|red_o|.

\begin{verbatim}
def distance_sweep(lights_state):
    return np.array(np.array([lights_state + [x] for 
                             x in np.linspace(0, 1, STEPS+1)]))

red_sweep = distance_sweep([1, 0, 0])
red_o = [np.rint(x[0]) for x in model.predict(red_sweep)]
\end{verbatim}

Once these input/output sweeps are generated for every possible light state, the ones where the partial differential $m$ is equal to 1 (modulo rounding). This gives all the critical points in the sweep where the output changes from 0 to 1, or 1 to 0. These are the points where the change in input results a dramatic change in the output. These input/output pairs are printed out by the probing stage in the input format of CNPInduce. Along with the input/output pairs, a distinct set of all the constants involved in the input/output pairs are extracted as well, and a synthesis job file is generated, as exemplified below:

\begin{verbatim}
jobValence([rd:in, am:in, gr:in, dist:in, go:out]).
%% Constants
jobConstant(0.0).
jobConstant(0.08).
jobConstant(0.09).
jobConstant(0.59).
jobConstant(0.6).
jobConstant(0.78).
jobConstant(0.79).
jobConstant(1.0).
%% Observables
jobObservable([rd:1.00, am:0.00, gr:0.00, dist:0.00, go:0.00], true).
jobObservable([rd:1.00, am:0.00, gr:0.00, dist:0.59, go:0.00], true).
jobObservable([rd:1.00, am:0.00, gr:0.00, dist:0.60, go:1.00], true).
jobObservable([rd:1.00, am:0.00, gr:0.00, dist:1.00, go:1.00], true).
jobObservable([rd:0.00, am:1.00, gr:0.00, dist:0.00, go:1.00], true).
jobObservable([rd:0.00, am:1.00, gr:0.00, dist:0.08, go:1.00], true).
jobObservable([rd:0.00, am:1.00, gr:0.00, dist:0.09, go:0.00], true).
jobObservable([rd:0.00, am:1.00, gr:0.00, dist:0.78, go:0.00], true).
jobObservable([rd:0.00, am:1.00, gr:0.00, dist:0.79, go:1.00], true).
jobObservable([rd:0.00, am:1.00, gr:0.00, dist:1.00, go:1.00], true).
jobObservable([rd:0.00, am:0.00, gr:1.00, dist:0.00, go:1.00], true).
jobObservable([rd:0.00, am:0.00, gr:1.00, dist:1.00, go:1.00], true).
\end{verbatim}

Each \verb|jobObservable| line in this synthesis file gives an input/output pair along with a Boolean flag \verb|true| that indicates if the model \textit{does} produce this output. Examples where the model \textit{does not} produce an output can alternatively be included for the synthesis to be able to eliminate some of the possible programs. 

\subsection{Synthesis: Synthesizing a human-readable program}
In the synthesis stage, the synthesis job file prepared by the previous stage is run through CNPInduce by loading it through the \verb|jobFromLocalFile| command. This command loads the constants and  examples (observables) from the file, and initiates a synthesis job with these. In this case, the number of domains in the model goes beyond the synthesizer's current efficiency limits. Therefore we manipulate the synthesis job file to only include the observables where the state of the light is red, which reduces the number of domains to 3, as shown:

\begin{verbatim}
jobObservable([rd:1.00, dist:0.00, go:0.00], true).
jobObservable([rd:1.00, dist:0.59, go:0.00], true).
jobObservable([rd:1.00, dist:0.60, go:1.00], true).
jobObservable([rd:1.00, dist:1.00, go:1.00], true).
\end{verbatim}

When the synthesizer is initiated with these observables, the first program it suggests is the following CNP program:
\begin{verbatim}
ande(const(rd,1.0),proj(iif(ltValue(a,0.6),0.0,1.0),[a->dist,o->go]))
\end{verbatim}

This reveals that the synthesizer found a program that involves all of the given arguments (rd, dist, go). CNPInduce also guarantees that the programs it produces are terminating programs. This means that as long as the inputs are constant values the programs generated will terminate and assign an output value. When a program is suggested, the synthesizer gives the option to stop or to look for other programs. When instructed to look for other programs, the synthesizer continues to find others that are correct, but longer and more complex. These induced logic programs suggested by CNPInduce form our explanation models.

In the next section, let us discuss how CNP programs can be interpreted as explanation models to validate the black box model being studied.

\subsection{Validation: Interpreting the program}

CNP is a pure relational language and therefore it may not be straightforward to those who are not familiar to such languages. In definite clause form it can be translated to:
\begin{align*}
& \text{model}(\text{Rd}, \text{Dist}, Go) \leftarrow  \text{Rd}=1.0 \wedge \text{Rd} < 0.6 \wedge \text{Go}=1.0  \\
& \text{model}(\text{Rd}, \text{Dist}, Go) \leftarrow  \text{Rd}=1.0 \wedge \neg(\text{Rd} < 0.6) \wedge \text{Go}=0.0 
\end{align*}

Or in English it can be written as:

\begin{displayquote}
\emph{If the red light is 1.0, when the Dist is less than 0.6 assign Go to 1.0, otherwise assign Go to 0.0.}
\end{displayquote}

\begin{figure}
\includegraphics[width=\textwidth]{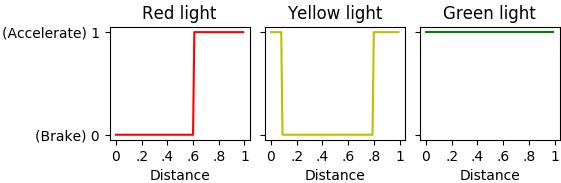}
\caption{Sensitivity analysis on the trained model, visualized.}
\label{fig_lights_plot}
\end{figure}

A visualization of the input/output states of the model can be seen in Figure \ref{fig_lights_plot}. The red line shows a shift at $Distance=0.6$, which is in line with the CNP program. Visualizations like these are useful but not always applicable due to higher dimensions or discrete data. In these cases a program is much more helpful to make sense of.

Once an explanation model expressed in CNP is available for a given black box model, it can be automatically translated to logical rules or to natural language such as structured English. Once in this form, further automated assertions can be made on the explanation model's CNP program. For example, we can use automated testing and validation on the CNP explanations to reason about regulatory and legal compliance, where law may be represented logical axioms \cite{Prakken2015LawAL}.

\section{Discussion and Future Work}
\label{discussion}
While we have demonstrated that RICE can be used to generate explanation models in this paper, we have identified a number of challenges that need to be addressed to improve its applicability, usability and performance.  

A weakness of our method lies in indirection. The probing stage extracts examples and these examples are used for program synthesis, which introduces a level of indirection between the synthesized program (the explanation model) and the actual model (the black box being studied). One way we suggest approaching this is to exploit the fact that both models can be evaluated for any input. By generating randomized samples for inputs and comparing the difference of the explanation model's output to those of the black box model, a difference value can be calculated that would converge to 0 if they are semantically identical. 

An opportunity that was discovered during the development of probing phase was that it was just as easy to identify erroneous behaviour in irregular system states (such as red light and green light being on at the same time), and specifically how the errors occur. As the probing stage indifferently extracts all critical data points, it identifies irregular states as well, which are reflected in the synthesized program. This is an opportunity that would be useful for identifying incorrect behaviour trained into ML models that might only use positive examples. Another point that needs improvement is the granularity of probing. In the demonstration presented in this paper, 1/100 granularity was used. Ideally the probing can be much more efficient using techniques such as logarithmic search and dynamic granularity. Using heuristics to make this stage more efficient is an acceptable approach since the final program can always be confirmed with randomized input/output pairs against the actual model.

An advantage of our generic probing stage is that the only manual requirement it needs is the valid value ranges of input/output variables, and a name for each so the synthesized program can refer to these names. The probing stage can be written in a completely generic fashion and since the synthesis stage is also domain-agnostic, the whole method can be considered generic.

We found that the success of synthesis is highly sensitive to the accuracy of the model itself. In the example, the model was measured to have over 99\% accuracy. With models under 90\% accuracy we did not find the synthesis successful in identifying a program. 

Scaling of the RICE method is a significant issue. Currently, the CNPInduce program synthesizer is implemented as a Prolog program, which cannot be parallelized trivially. But the simplicity of the search algorithm behind CNPInduce \cite{Pacaci2017Representation} may allow for a MapReduce implementation that is more amenable to parallelization \cite{Chu2007MapReduceForML}. 

Finally, while the usability of CNP has been studied \cite{Pacaci2017Representation}, its use as a target language for explanation models needs to be tested. Studies have shown that model interpretability (transparent models vs. explained black box models) and the styles of algorithmic explanations have varying degrees to which they build trust in ML models and there is currently no `best' approach emerging \cite{DBLP:journals/corr/abs-1802-07810,Binns:2018:RHP:3173574.3173951}.

\section{Conclusions}
\label{conc}

In this paper we have demonstrated a novel method of assigning meaning to opaque software artifacts whose specification is unknown. Our methodology called RICE, does not try to interfere, steer or model the black box directly. Instead we have adopted a three stage post-processing approach in which we use the artefact as an opaque prototype on which to extract meaning. Our first stage, called \textit{probing}, inspects the generated artefact using sensitivity analysis to create a set of valid input/output pairs. From these pairs we have shown how to generate a logic-based explanation model, using \textit{synthesis} as a second stage, that subsequently allows the artefact to be interpreted. This is achieved through a final stage, called \textit{validation}, that produces a readable version.

This paper presents a proof of concept and much work remains to successfully scale the technique up to larger and more complex examples. Our interests lie not in the popular areas of machine learning, such as image recognition and consumer marketing, but in applications that require legal interpretation and certification. 

Finally while the interoperability problem is usually presented as a concern for systems based on machine learning, the scope is far greater. Legacy systems for which there is no longer a readable source code, or agent based systems that have been finely calibrated are two further examples that would benefit from the ability to reverse-engineer an explainable interpretation that our methodology permits.

\bibliographystyle{acm}  
\bibliography{references}       
\end{document}